\documentclass{article}
\usepackage{spconf}
\usepackage{amsmath,amssymb,bm,graphicx,booktabs}
\usepackage{tikz}
\usetikzlibrary{arrows.meta,positioning,calc}

\title{Task-Directed Residual AddUNet:\\
Perfect-Reconstruction Routing for Full-Rate Representations}

\name{Vikram R. Lakkavalli}
\address{International Institute of Information Technology Bangalore, India\\
\texttt{vikram.ckm@gmail.com, vikram.ramesh@iiitb.ac.in}}

\begin{document}
\ninept
\maketitle

\begin{abstract}
This paper establishes a perfect-reconstruction (PR) interpretation of
AddUNet and introduces a Residual Full-Rate PR architecture for task-directed
representation learning. The survivor--skip structure of a constrained
additive U-Net is shown to be exactly equivalent to a critically sampled
multirate PR filter bank. Removing critical sampling yields a full-rate PR
setting free from decimation-induced alias-cancellation constraints. Within
this setting, Residual Full-Rate PR progressively routes structure away from a
task-facing survivor while retaining the routed information explicitly. Exact
reconstruction is guaranteed for arbitrary shape-compatible linear or
nonlinear routing operators, without requiring invertibility, a matched
synthesis bank, reconstruction loss, or a learned decoder. The resulting
architecture decouples representation design from reconstruction design:
conservation is structural, while learning is devoted to routing. The same
formulation reveals that an identity-shortcut ResNet block becomes a full-rate
PR stage when its residual output is retained; recovery then follows from the
retained complement rather than from invertibility of the block map. This
principle is verified on a pretrained ResNet-34. On TIMIT, the proposed
front-end improves test PER from $28.60\pm2.09\%$ to $25.76\pm0.41\%$ with
the recognizer and training protocol held fixed, while maintaining exact
reconstruction. Speaker probing further shows that structural conservation
does not itself imply task-specific invariance.
\end{abstract}

\begin{keywords}
perfect reconstruction, residual learning, skip connections, additive U-Net,
residual networks, task-directed routing
\end{keywords}

\section{Introduction}

Encoder--decoder networks with skip connections are widely used in signal and
image tasks, yet the skip is typically treated as an engineering mechanism for
passing detail around a bottleneck, leaving the signal-theoretic role of the
individual paths implicit~\cite{unet}. For task-directed representation
learning, however, it is useful to distinguish what information is propagated
from what is explicitly routed away.

Additive U-Nets provide a structured case because the propagated survivor and
the retained skip are recombined by addition at synthesis. This paper shows
that, under appropriate constraints, this organization is exactly equivalent
to a critically sampled multirate perfect-reconstruction (PR)
analysis--synthesis filter bank~\cite{vaidyanathan,vetterli}. The result places
the survivor--skip architecture within classical multirate theory and provides
an explicit account of how information is conserved across the two paths.
Related learnable multirate systems such as DeSpaWN realize PR principles in
wavelet-style analysis~\cite{despawn}; the concern here is the generic
survivor--skip structure underlying AddUNet~\cite{addunet}.

Critical sampling, however, also imports the constraints of the multirate
setting. Analysis and synthesis must jointly satisfy the PR conditions in the
presence of decimation, including alias cancellation. Removing critical
sampling opens a full-rate PR setting in which the two analysis paths need not
form complementary spectral subbands, although analysis and synthesis remain
jointly constrained by PR.

Within this full-rate setting, this paper introduces a \emph{Residual Full-Rate
PR} architecture for task-directed representation learning. Each stage routes
a component away from the propagated survivor and retains that component
explicitly. Exact reconstruction then follows structurally by addition, so the
routing operator may be selected or learned for the representation task rather
than for synthesis. This decouples information conservation from information
organization: conservation is fixed by architecture, while learning determines
what remains in the task-facing survivor and what is transferred to the
retained paths.

The same formulation exposes a broader connection to residual networks. For an
identity-shortcut ResNet block, retaining the usual residual output makes the
block exactly recoverable by additive synthesis, even though the block mapping
itself need not be invertible. This is distinct from i-RevNet, i-ResNet, and
invertible U-Nets, which obtain recoverability by constraining the
transformation itself~\cite{jacobsen2018irevnet,behrmann2019iresnet,
etmann2020iunet}. Here, recoverability follows from retaining complementary
side information, trading transform constraints for representation redundancy.

The paper makes two main theoretical contributions. First, it establishes that
the constrained additive U-Net is exactly a critically sampled multirate PR
filter bank and shows that removing critical sampling yields a full-rate PR
design space free from decimation-induced alias-cancellation constraints.
Second, it introduces Residual Full-Rate PR and proves that exact
reconstruction holds for arbitrary shape-compatible linear or nonlinear
routing operators without an invertible transform, matched synthesis bank,
reconstruction loss, or learned decoder. A direct consequence is a retained-
residual PR interpretation of identity-shortcut ResNet blocks.

The experiments provide supporting findings. The retained-residual result is
verified on identity-shortcut blocks of an ImageNet-pretrained ResNet-34. On
TIMIT phone recognition, the Residual Full-Rate PR front-end improves
recognition over direct input under the same recognizer and training protocol
while preserving exact reconstruction. Routing width and local context show
non-monotonic effects, and speaker probing shows that structural conservation
alone does not imply task-specific invariance.

\section{Multirate PR Equivalence and Residual Full-Rate PR}
\label{sec:theory}

The architectural form considered here retains the encoder--decoder topology
and skip structure of an additive U-Net. At level $l$, let $p_{l-1}$ denote the
level input, $p_l$ the propagated survivor, and $q_l$ the retained skip
contribution.

\subsection{Critically Sampled PR}

In a critically sampled realization, $p_l$ and $q_l$ are the two decimated
outputs of a two-channel analysis bank. Let $\mathbf{E}_l(z)$ and
$\mathbf{R}_l(z)$ denote the corresponding analysis and synthesis polyphase
matrices. Then
\begin{equation}
\begin{bmatrix}p_l\\q_l\end{bmatrix}
=\mathbf{E}_l(z)
\begin{bmatrix}(p_{l-1})_e\\(p_{l-1})_o\end{bmatrix},
\label{eq:analysis_polyphase}
\end{equation}
with synthesis
\begin{equation}
\begin{bmatrix}(\hat p_{l-1})_e\\(\hat p_{l-1})_o\end{bmatrix}
=\mathbf{R}_l(z)
\begin{bmatrix}p_l\\q_l\end{bmatrix}.
\label{eq:synthesis_polyphase}
\end{equation}
Perfect reconstruction holds whenever
\begin{equation}
\mathbf{R}_l(z)\mathbf{E}_l(z)=z^{-\Delta_l}\mathbf{I},
\label{eq:pr_condition}
\end{equation}
which reconstructs $p_{l-1}$ exactly up to delay~\cite{vaidyanathan,vetterli}.
For a paraunitary realization,
\begin{equation}
\mathbf{E}_l^{H}(z^{-1})\mathbf{E}_l(z)=\mathbf{I}.
\label{eq:paraunitary}
\end{equation}
Thus the constrained additive multirate U-Net is a two-channel critically
sampled PR filter bank. More generally, decimation couples the analysis and
synthesis branches through alias-cancellation and distortionless-
reconstruction conditions; the two paths therefore cannot be designed
independently solely according to a representation objective.

\subsection{Full-Rate Generalization}

In a critically sampled two-channel filter bank, the input is represented
through a specific rate-preserving polyphase decomposition, conventionally
associated with its even and odd components. Perfect reconstruction requires
that these two branches jointly reproduce the signal that generated them.

Once critical sampling is no longer required, the same principle can be
retained without restricting the decomposition to the classical polyphase
form. At level $l$, let
\begin{equation}
p_l=P_l p_{l-1}, \qquad q_l=Q_l p_{l-1},
\label{eq:fullrate_analysis}
\end{equation}
where both components remain at the original rate. A linear synthesis stage
\begin{equation}
\hat p_{l-1}
=
\widetilde P_l p_l+\widetilde Q_l q_l
\label{eq:fullrate_synthesis}
\end{equation}
is perfectly reconstructing whenever
\begin{equation}
\widetilde P_lP_l+\widetilde Q_lQ_l=\mathbf I.
\label{eq:fullrate_pr}
\end{equation}

Thus, the essential requirement is not the particular even--odd
decomposition, but that each pair $(p_l,q_l)$ jointly reproduces its parent
$p_{l-1}$. If this condition is satisfied at every level, the complete
hierarchy reconstructs the original input exactly. The individual branches
may redistribute or reorganize signal content, while their joint synthesis
preserves the input without distortion.

This full-rate formulation therefore enlarges the admissible decomposition
space: the two branches need not form a classical complementary-subband or
alias-cancelling pair, provided that their joint representation satisfies
\eqref{eq:fullrate_pr}.

\noindent\textbf{Proposition 1 (Hierarchical full-rate PR).}
Consider an $L$-level hierarchy in which each level produces
$(p_l,q_l)$ from $p_{l-1}$ and admits a synthesis operation
$\mathcal S_l$ satisfying
\begin{equation}
\mathcal S_l(p_l,q_l)=p_{l-1}.
\end{equation}
Then recursive synthesis from level $L$ reconstructs the original input
$p_0$ exactly.

\noindent\emph{Proof.}
Since $(p_L,q_L)$ reconstructs $p_{L-1}$, the recovered $p_{L-1}$
together with $q_{L-1}$ reconstructs $p_{L-2}$. Repeating this argument
recursively yields $p_0$.\hfill$\square$.

Thus, perfect reconstruction of the hierarchy is a local property: each level
need only preserve sufficient complementary information to reproduce the
representation that generated it. The internal survivor and routed paths may
reorganize the signal arbitrarily, but their joint synthesis must recover the
parent representation exactly. Consequently, the complete hierarchy is
transparent to the input; for a linear system, exact reconstruction also
implies preservation of the input spectrum end-to-end.

\subsection{Proposed Residual Full-Rate PR}
\label{sec:residual}

\subsection{Residual Full-Rate PR}

The preceding condition admits many possible full-rate PR realizations. We
propose a particularly simple one suited to task-directed representation
learning. Let the routed component be
\begin{equation}
q_l=Q_l(p_{l-1}),
\end{equation}
and define the survivor as its explicit complement,
\begin{equation}
p_l=p_{l-1}-q_l.
\label{eq:residual_analysis}
\end{equation}
The parent representation is then recovered by the fixed synthesis rule
\begin{equation}
p_{l-1}=p_l+q_l.
\label{eq:residual_synthesis}
\end{equation}

For linear $Q_l$, this corresponds to
\begin{equation}
P_l=\mathbf I-Q_l,
\qquad
\widetilde P_l=\widetilde Q_l=\mathbf I,
\end{equation}
and therefore
\begin{equation}
\widetilde P_lP_l+\widetilde Q_lQ_l
=
(\mathbf I-Q_l)+Q_l
=
\mathbf I.
\end{equation}
Hence the proposed residual construction is one realization of the general
full-rate PR condition in \eqref{eq:fullrate_pr}.

Its key advantage is that the PR constraint no longer restricts the routing
operator itself. The operator $Q_l$ determines what is routed away from the
survivor, while exact reconstruction follows structurally from retaining the
routed component. This allows the hierarchy to devote its learned capacity to
representation design rather than synthesis design.
\begin{figure}[t]
\centering
\resizebox{\columnwidth}{!}{%
\begin{tikzpicture}[
    >=Latex,
    font=\scriptsize,
    line width=0.75pt,
    box/.style={draw,rounded corners=2pt,align=center,minimum height=1.05cm},
    io/.style={box,minimum width=1.65cm},
    stage/.style={box,fill=blue!10,minimum width=3.05cm,minimum height=1.45cm},
    sum/.style={draw,circle,inner sep=0pt,minimum size=6.5mm,font=\normalsize},
    arr/.style={->},
    darr/.style={->,dashed}
]
\node[anchor=east,font=\bfseries] at (-0.15,0.35) {Analysis};
\node[io]    (in)   at (1.35,0)   {input\\$x=p_0$};
\node[stage] (st1)  at (4.65,0)   {{\bfseries Stage 1}\\$q_1=Q_1(p_0)$\\$p_1=p_0-q_1$};
\node                at (7.35,0)   {$\cdots$};
\node[stage] (stL)  at (10.05,0)  {{\bfseries Stage $L$}\\$q_L=Q_L(p_{L-1})$\\$p_L=p_{L-1}-q_L$};
\node[io]    (lat)  at (13.2,0)   {latent\\$p_L$};
\draw[arr] (in.east)--(st1.west);
\draw[arr] (st1.east)--(6.45,0);
\draw[arr] (8.25,0)--(stL.west);
\draw[arr] (stL.east)--(lat.west);

\node[anchor=east,font=\bfseries] at (-0.15,-2.45) {Synthesis};
\node[io]  (rec)   at (1.35,-2.75)  {reconstructed\\$\hat x=\hat p_0$};
\node[sum] (sum1)  at (4.65,-2.75)  {$+$};
\node               at (7.35,-2.75)  {$\cdots$};
\node[sum] (sumL)  at (10.05,-2.75) {$+$};
\node[io]  (start) at (13.2,-2.75)  {start\\$\hat p_L=p_L$};
\draw[arr] (start.west)--(sumL.east);
\draw[arr] (sumL.west)--(8.25,-2.75);
\draw[arr] (6.45,-2.75)--(sum1.east);
\draw[arr] (sum1.west)--(rec.east);
\draw[darr] (st1.south)--node[right] {$q_1$}(sum1.north);
\draw[darr] (stL.south)--node[right] {$q_L$}(sumL.north);
\draw[arr] (lat.south)--(start.north);
\end{tikzpicture}%
}
\caption{Residual Full-Rate PR AddUNet. Each stage routes
$q_l=Q_l(p_{l-1})$, propagates $p_l=p_{l-1}-q_l$, and reconstructs in reverse
by fixed addition.}
\label{fig:residual_pr_addunet}
\end{figure}
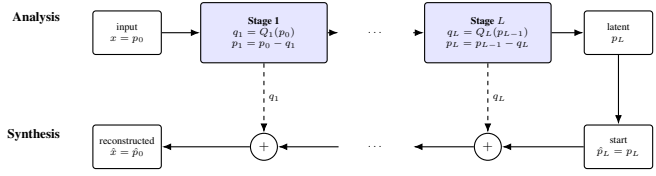

\subsection{Residual Networks as Retained-Residual PR Stages}
\label{sec:resnet_theory}

Consider the standard identity-shortcut residual update~\cite{he2016resnet}
\begin{equation}
p_l=p_{l-1}+F_l(p_{l-1}).
\label{eq:resnet_update}
\end{equation}
Let
\begin{equation}
r_l=F_l(p_{l-1})
\label{eq:resnet_residual}
\end{equation}
be retained during the forward pass. Then the preceding state is recovered
exactly as
\begin{equation}
p_{l-1}=p_l-r_l.
\label{eq:resnet_synthesis}
\end{equation}
Thus an identity-shortcut residual block, together with its retained residual,
forms a redundant full-rate PR analysis--synthesis stage. Under the convention
of \eqref{eq:residual_analysis}, this is the same construction with
$Q_l=-F_l$ and $q_l=-r_l$.

\noindent\textbf{Corollary 1 (Identity-shortcut residual cascade).}
\emph{For a sequence of shape-preserving identity-shortcut residual blocks,
retaining every residual $r_l$ makes the joint representation
$(p_L,r_1,\ldots,r_L)$ exactly recoverable as}
\begin{equation}
p_0=p_L-\sum_{l=1}^{L}r_l.
\label{eq:resnet_cascade}
\end{equation}
The result follows directly by telescoping
$p_l=p_{l-1}+r_l$.

This result is a recoverability statement about the \emph{joint}
representation, not an invertibility claim about the block map
$p_{l-1}\mapsto p_l$. In i-RevNet, i-ResNet, and invertible U-Nets,
recoverability is obtained by constraining the transformation itself to admit an
inverse~\cite{jacobsen2018irevnet,behrmann2019iresnet,etmann2020iunet}. In the
retained-residual construction, no such condition is needed: the missing
complement is carried as side information. Projection or downsampling blocks
are outside this identity-shortcut result unless their lost information is also
retained.

\section{Experiments}
\label{sec:experiments}

\subsection{Architectural Verification on a Pretrained ResNet}

To verify that the retained-residual result is architectural rather than
specific to AddUNet or speech, we apply \eqref{eq:resnet_synthesis} directly to
the three identity-shortcut blocks of \texttt{layer1} in an ImageNet-pretrained
ResNet-34~\cite{he2016resnet}. The network and forward computation are left
unchanged; only the per-block residual
$r_l=F_l(p_{l-1})$ is retained. Reverse synthesis then applies
$p_{l-1}=p_l-r_l$ block by block, with no retraining or invertibility
constraint.

Relative $\ell_2$ reconstruction error remains approximately $10^{-8}$ in
single precision and $10^{-17}$ in double precision across depth. The result
therefore verifies that exact recoverability follows from retained residual
information rather than invertibility of the pretrained block. Projection and
downsampling transitions are not included in this experiment.

\subsection{TIMIT Setup}

We evaluate Residual Full-Rate PR AddUNet on TIMIT using 257-dimensional
log-magnitude STFT features and the standard 61-to-39 phone mapping. The
full-rate hierarchy has six routing stages with temporal dilations
$\{1,1,2,2,4,4\}$; each $Q_l$ contains two time--frequency convolutions with
batch normalization and ReLU between them. For $C$ channels, the input is
redundantly lifted to $C$ full-rate channels and collapsed before a common CTC
backend: CMVN, a 257-to-128 projection, two BiLSTM layers with 128 hidden
units, layer normalization, and a 40-way classifier.

Models use CTC only, batch size 8, learning rate $2\times10^{-4}$, weight decay
$10^{-5}$, and seeds $\{1234,1235,1236\}$. Checkpoints are selected by
validation PER. No reconstruction loss is used. Relative reconstruction error
remains below $6.4\times10^{-8}$ in all PR runs.

As a controlled baseline, the complete routing hierarchy is bypassed and the
original log-magnitude spectrum is fed directly to the same CTC backend. Thus
the direct-input and PR models share the recognizer, optimizer, data splits,
and model-selection procedure; the only difference is whether the task
representation is the original input or the routed survivor $p_L$.

\subsection{Routing Capacity and Context}

We vary routing width $C\in\{1,2,4,8,12\}$ with a $9\times3$ kernel, then fix
$C=2$ and vary spectral context using $9\times3$, $9\times5$, and $9\times9$
kernels; $C=1,9\times9$ is included as a single-channel control.

\begin{table}[t]
\centering
\caption{TIMIT routing-capacity and context ablations; mean $\pm$ std. over
three seeds.}
\label{tab:routing_ablation}
\footnotesize
\begin{tabular}{@{}ccrr@{}}
\toprule
$C$ & Kernel & Val PER (\%) & Test PER (\%) \\
\midrule
1  & $9\times3$ & $24.34\pm1.68$ & $26.68\pm1.43$ \\
2  & $9\times3$ & $23.59\pm0.81$ & $26.03\pm0.95$ \\
4  & $9\times3$ & $24.61\pm1.75$ & $27.12\pm1.53$ \\
8  & $9\times3$ & $24.30\pm0.45$ & $26.48\pm0.44$ \\
12 & $9\times3$ & $23.92\pm1.36$ & $26.31\pm1.28$ \\
\midrule
1  & $9\times9$ & $24.17\pm1.44$ & $26.56\pm1.59$ \\
\textbf{2} & $\mathbf{9\times5}$ & $\mathbf{23.40\pm0.45}$ & $\mathbf{25.76\pm0.41}$ \\
2  & $9\times9$ & $24.13\pm0.34$ & $26.23\pm0.20$ \\
\bottomrule
\end{tabular}
\end{table}

A single routing channel is already effective, while $C=2$ gives the best mean
recognition among the $9\times3$ models. Further width gives no monotonic gain.
At fixed $C=2$, moderate spectral context ($9\times5$) gives the best overall
test PER, whereas $9\times9$ provides no further benefit. Thus routing
capacity and local context affect the learned representation differently, and
neither improves monotonically with scale.

\subsection{Controlled Baseline and AddUNet Reference}

\begin{table}[t]
\centering
\caption{TIMIT 39-phone test PER. Direct CTC and PR-AddUNet use the same
classifier backend and training protocol and are averaged over three seeds.
AddUNet results are published five-seed reference points.}
\label{tab:baseline}
\footnotesize
\begin{tabular}{@{}lcr@{}}
\toprule
Model & Objective & Test PER (\%) \\
\midrule
Direct input + CTC & CTC & $28.60\pm2.09$ \\
PR-AddUNet ($C{=}2$, $9{\times}5$) & CTC, exact PR & $25.76\pm0.41$ \\
Real AddUNet~\cite{addunet}   & CTC+recon. & $23.97\pm1.18$ \\
Pseudo AddUNet~\cite{addunet} & CTC+recon. & $23.30\pm1.25$ \\
\bottomrule
\end{tabular}
\end{table}

With the same classifier and training protocol, adding the Residual Full-Rate
PR front-end improves mean validation PER from $26.70\pm2.20\%$ to
$23.40\pm0.45\%$ and mean test PER from $28.60\pm2.09\%$ to
$25.76\pm0.41\%$, a 2.84-point absolute test improvement, while retaining
exact reconstruction by construction. Because the controlled baseline removes
the front-end entirely, this result does not isolate the contribution of the
residual/PR structure from the benefit of a generic learned front-end; a
capacity-matched non-PR control is left to future work.

The earlier AddUNet~\cite{addunet} remains about two PER points better, but it
uses a lossy bottleneck and joint CTC-plus-reconstruction training and is
therefore treated only as an external reference rather than a controlled
ablation.

\subsection{Speaker Information in the Survivor}

We freeze the ASR models and train linear speaker probes on utterance-level
mean and standard-deviation statistics of the input and final survivor. Input
top-1 accuracy is $76.30\%$; for $C=\{1,2,4,8,12\}$, survivor accuracies are
$75.18\pm2.37\%$, $75.90\pm0.74\%$, $73.52\pm1.66\%$, $73.88\pm1.11\%$,
and $73.41\pm1.47\%$, respectively. Thus phone supervision alone reduces
linear speaker accessibility by at most about three points.

By comparison, the lossy AddUNet reduces its own input probe from $78.97\%$ to
$60.60\%$ at the bottleneck~\cite{addunet}; because the probe protocols differ,
only within-model drops are compared. The weaker suppression observed here is
not imposed by PR: exact reconstruction constrains the joint survivor--residual
representation, not the survivor alone. Nuisance information could in
principle be routed predominantly to the retained residual paths, but phone
supervision alone does not strongly induce that organization.

\section{Discussion}

The PR interpretation gives the U-Net hierarchy a role beyond information
bypass. At each level, the representation is divided into a retained routed
component $q_l$ and a propagated survivor $p_l$. In Residual Full-Rate PR,
$Q_l$ may be any shape-compatible nonlinear operator while
$p_{l-1}=p_l+q_l$ keeps every routing decision exactly recoverable. The
architecture therefore converts representation learning into a sequence of
routing decisions under a fixed conservation law.

The ResNet consequence shows that this retained-complement view is not tied to
the U-Net topology or to speech. A standard identity-shortcut residual block
already has the required algebraic structure; retaining its residual simply
makes the complementary variable explicit. The key distinction from
invertible residual networks is therefore where recoverability is placed: in
the transform itself for invertible networks, versus in the redundant joint
representation for retained-residual PR.

For task-directed learning, the architecture of $Q_l$ controls what can be
routed at each level. Channel width determines routing capacity, while depth
and receptive field determine the nonlinearity and context available to each
stage. Because successive stages operate on a progressively conditioned
survivor, these requirements need not be uniform across depth. The observed
non-monotonic effects of width and context are consistent with this view:
additional capacity does not necessarily produce a more useful decomposition.

This also clarifies the earlier AddUNet results~\cite{addunet}. PR guarantees
conservation of the joint survivor--residual representation but does not decide
which factors remain in the survivor. The stronger reduction in linear speaker
accessibility previously obtained with a lossy bottleneck and joint
recognition--reconstruction training is therefore consistent with the
interaction of bottleneck capacity, reconstruction objective, nonlinear
routing, and phone supervision rather than additive topology alone.

\section{Conclusion}

This paper establishes a PR interpretation of additive U-Nets and introduces
Residual Full-Rate PR as a task-directed architecture in which exact
reconstruction is structural. By retaining the routed complement, arbitrary
shape-compatible linear or nonlinear routing operators can be used without an
invertible transform, matched synthesis bank, reconstruction loss, or learned
decoder. The same retained-residual principle reveals identity-shortcut ResNet
blocks as exactly recoverable full-rate PR stages when their residual outputs
are retained.

On TIMIT, the resulting routing front-end improves test PER from
$28.60\pm2.09\%$ to $25.76\pm0.41\%$ under the same recognizer and training
protocol while preserving exact reconstruction. Routing width and local
context exhibit non-monotonic effects, and speaker probing shows that PR
guarantees conservation but not task-specific invariance. Future work will
study explicit routing constraints, level-dependent operator design, and
controlled departures from PR through compressed representations requiring
learned synthesis.


\begin{thebibliography}{99}

\bibitem{unet}
O.~Ronneberger, P.~Fischer, and T.~Brox,
``U-Net: Convolutional networks for biomedical image segmentation,''
in \emph{Proc. MICCAI}, 2015, pp.~234--241.

\bibitem{vaidyanathan}
P.~P. Vaidyanathan,
\emph{Multirate Systems and Filter Banks}.
Englewood Cliffs, NJ, USA: Prentice-Hall, 1993.

\bibitem{vetterli}
M.~Vetterli and J.~Kova\v{c}evi\'c,
\emph{Wavelets and Subband Coding}.
Englewood Cliffs, NJ, USA: Prentice-Hall, 1995.

\bibitem{despawn}
G.~Michau, G.~Frusque, and O.~Fink,
``Fully learnable deep wavelet transform for unsupervised monitoring of
high-frequency time series,''
\emph{Proc. Natl. Acad. Sci. USA}, vol.~119, no.~8,
Art.~no.~e2106598119, 2022.

\bibitem{addunet}
V.~R. Lakkavalli and N.~Sinha,
``Architectural control of phonetic invariance via additive U-Net
multi-task learning,''
in \emph{Proc. Int. Conf. Signal Processing and Communications (SPCOM)},
2026, pp.~1--5, doi: 10.1109/SPCOM71105.2026.11622998.

\bibitem{he2016resnet}
K.~He, X.~Zhang, S.~Ren, and J.~Sun,
``Deep residual learning for image recognition,''
in \emph{Proc. IEEE Conf. Computer Vision and Pattern Recognition (CVPR)},
2016, pp.~770--778.

\bibitem{jacobsen2018irevnet}
J.-H.~Jacobsen, A.~W.~M.~Smeulders, and E.~Oyallon,
``i-RevNet: Deep invertible networks,''
in \emph{Proc. Int. Conf. Learning Representations (ICLR)}, 2018.

\bibitem{behrmann2019iresnet}
J.~Behrmann, W.~Grathwohl, R.~T.~Q.~Chen, D.~Duvenaud, and J.-H.~Jacobsen,
``Invertible residual networks,''
in \emph{Proc. 36th Int. Conf. Machine Learning (ICML)}, vol.~97,
2019, pp.~573--582.

\bibitem{etmann2020iunet}
C.~Etmann, R.~Ke, and C.-B.~Sch{\"o}nlieb,
``iUNets: Learnable invertible up- and downsampling for large-scale inverse problems,''
in \emph{Proc. IEEE 30th Int. Workshop Machine Learning for Signal Processing (MLSP)},
2020, pp.~1--6.

\end{thebibliography}
\end{document}